\documentclass{article} 
\usepackage{colm2024_conference}

\usepackage{microtype}
\usepackage{hyperref}
\usepackage{url}
\usepackage{graphicx}
\usepackage{booktabs}
\definecolor{darkblue}{rgb}{0, 0, 0.5}
\hypersetup{colorlinks=true, citecolor=darkblue, linkcolor=darkblue, urlcolor=darkblue}

\title{CopilotCAD: Empowering Radiologists with Report Completion Models and Quantitative Evidence from Medical Image Foundation Models}

\author{Sheng Wang*,Tianming Du\thanks{Authors contributed equally to this work}\\
University of Pennsylvania\\
\texttt{\{sheng.wang, tinaming.du\}@pennmedicine.upenn.edu} 
\AND  Katherine Fischer, Gregory E Tasian\\
The Children’s Hospital of Philadelphia\\
\texttt{\{FISCHERK, TasianG\}@chop.edu} 
\AND
Justin Ziemba, Joanie M Garratt, Hersh Sagreiya, Yong Fan \\
University of Pennsylvania\\
\texttt{\{Justin.Ziemba, Joanie.Garratt, Hersh.Sagreiya, yong.fan\}@pennmedicine.upenn.edu}} 

\colmfinalcopy 
\begin{document}

\maketitle

\begin{abstract}
Computer-aided diagnosis systems hold great promise to aid radiologists and clinicians in radiological clinical practice and enhance diagnostic accuracy and efficiency. However, the conventional systems primarily focus on delivering diagnostic results through text report generation or medical image classification, positioning them as standalone decision-makers rather than helpers and ignoring radiologists' expertise. This study introduces an innovative paradigm to create an assistive co-pilot system for empowering radiologists by leveraging Large Language Models (LLMs) and medical image analysis tools. Specifically, we develop a collaborative framework to integrate LLMs and quantitative medical image analysis results generated by foundation models with radiologists in the loop, achieving efficient and safe generation of radiology reports and effective utilization of computational power of AI and the expertise of medical professionals. This 
approach empowers radiologists to generate more precise and detailed diagnostic reports, enhancing patient outcomes while reducing the burnout of clinicians. Our methodology underscores the potential of AI as a supportive tool in medical diagnostics, promoting a harmonious integration of technology and human expertise to advance the field of radiology.
\end{abstract}

\section{Introduction}

Medical imaging is a cornerstone of modern diagnostic procedures, and the advent of Computer-Aided Diagnosis (CAD) systems powered by Artificial Intelligence (AI) has been transformative. Initially developed to augment the accuracy and efficiency of image classification, these systems are now increasingly capable of generating detailed radiology reports, facilitated by recent advancements in Large Language Models (LLMs). The integration of LLMs into medical image analysis has the potential to revolutionize the radiology report generation by providing nuanced insights that go beyond the capabilities of traditional image classification models.

Despite their sophistication, current CAD systems face significant challenges, particularly in the realm of diagnostic results' explainability, transparency, and safety since their output is typically generated by a machine learning model without any supervision of human medical experts. The lack of cross-communication between CAD systems and human experts renders ineffective utilization of the intelligence power of both AI systems and human experts. Moreover, the opaqueness of typical machine learning models of CAD systems also hampers trust in CAD systems, commonly happened in all AI applications. Therefore, it is desired to develop a CAD system to best utilize AI and human experts' domain knowledge, empower clinical decision-making, and enhance trust in AI-supported systems.

Like clinical decision-making, other hard tasks that require expertise such as programming still cannot be fully automatic. However, semi-automated code completion tools, such as Github's Copilot, have gained widespread adoption. Additionally, other code completion large language models (LLMs), including Wizard Coder\citep{luo2023wizardcoder}, Stable Coder \citep{StableCodeInstructAlpha}, and CodeLlama\citep{rozière2024code}, are facilitating more rapid and reliable coding practices. These works utilize AI and human programmers' domain knowledge evenly.

Inspired by the Copilot and its related works, our approach introduces a hybrid CAD system, entitled CopilotCAD, to keep the human experts in the loop. As illustrated in Figure 1, CopilotCAD serves as a bridge between the conventional workflow and a fully automatic CAD. Particularly, our system enhances the radiology diagnostic process by integrating the computational efficiency of AI, implemented through LLMs and medical image foundation models, and the irreplaceable judgment of a radiologist. The system provides an friendly interface to facilitate interactive image based diagnosis, enabling radiologists to make informed decisions supported by AI-generated quantitative data and visual aids, such as volumetric measurements, informative radiomic features, and anomaly detection.

To substantiate our concept, we conducted an experiment to utilize CopilotCAD in a clinical setting. The preliminary results demonstrate a notable improvement in diagnostic accuracy and efficiency, empowering radiologists working with CopilotCAD to generate more comprehensive reports at a faster rate compared to using conventional methods. Moreover, the cognitive load on clinicians was reduced, indicating the system's potential to alleviate burnout of radiologists.
\begin{figure}[t]
\begin{center}
\label{fig:overview}

    \includegraphics[width=0.95\linewidth]{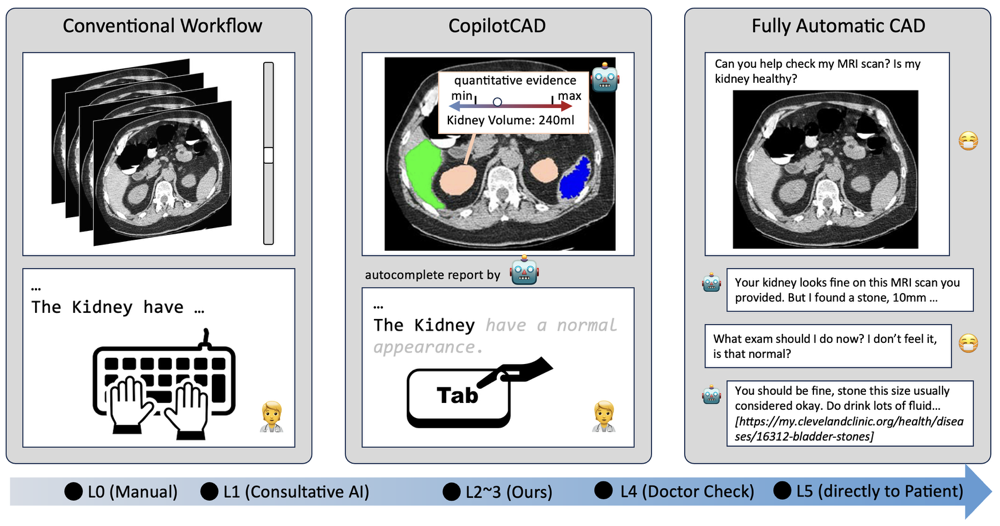}

\end{center}
\caption{CopilotCAD integrates the computational efficiency of AI and expertise of radiologists and provides an friendly interface to facilitate interactive image based diagnosis, enabling radiologists to make informed decisions supported by AI-generated quantitative data and visual aids with enhanced explainability, transparency, and safety, reflecting a paradigm shift away from traditional CAD systems.}
\end{figure}

The main contributions of this paper are summarized below:

\begin{itemize}
\item \textbf{Establishing an Assistive Role for CAD Systems}: We introduce a framework to complement the radiologist's workflow with computer-aided diagnosis systems by keeping the human experts in the decision-making loop and enhancing cross-communication between CAD systems and human experts.

\item \textbf{Optimizing AI Assistance Through LLMs and Medical Image Foundation Models}: The radiologist-AI interaction is enhanced by leveraging an integration of LLMs and medical image foundation models, with the latter generating quantitative measures of medical imaging data as \textit{\textbf{informative prompt}} to guide the LLMs. The medical image foundation models also generate visual aids to help with human experts' decision-making in addition to the quantitative imaging measures, providing concrete improvements in diagnostic precision while maintaining the radiologist's central role in interpretation.


\item \textbf{Promising preliminary results}: The new system has been validated from perspectives of radiology reporting quality (report completion) , demonstrating the new system's potential qualitatively and quantitatively with illustrative examples.

\end{itemize}

\section{Automatic Diagnosis Systems}


As illustrated in Figure \ref{fig:overview}, drawing an analogy to autonomous driving, radiological diagnosis systems can be categorized by their level of autonomy. Presently, the majority of systems function at Level 1 (L1) autonomy, providing preliminary results for radiologists to review and interpret. These systems, similar to basic driver-assistance technologies, augment the radiologist's capabilities by serving as a consultative tool rather than offering definitive diagnoses.

Advancements in AI have led to the development of systems that approach Level 4 (L4) autonomy, where AI delivers detailed diagnostic suggestions that medical expert may refine or validate. This stage resembles vehicles capable of navigating autonomously under certain conditions. Within this context, systems such as Med-PaLM, Med-PaLM 2~\citep{singhal2022large} and LLaVa-Med~\citep{li2023llavamed}, RadFM~\cite{wu2023towards} exemplify the L4 category by enabling interactive computer-aided diagnosis that requires radiologist oversight. Furthermore, a nascent category of diagnostic systems aligns with Level 5 (L5) autonomy, where AI assumes a more independent role by directly, searching on the internet, conveying diagnostic conclusions to patients, and chat with them to give further medical advice~\citep{wang2023chatcad,zhao2023chatcad+}, thus minimizing or even eliminating the need for radiologist intervention.


Contrary to the trend towards full autonomy, our proposed framework targets the integration of AI within the L2 to L3 autonomy spectrum. This model fosters a collaborative environment where AI supports the radiologist's diagnostic process, enhancing precision and reducing workload. By aligning with L2-L3 autonomy, our approach emphasizes the AI's role as an assistive co-pilot, underscoring the invaluable synergy between computational intelligence and human expertise in advancing radiological diagnostics.

\section{Approach}

\begin{figure}[h]
\begin{center}

    \includegraphics[width=0.95\linewidth]{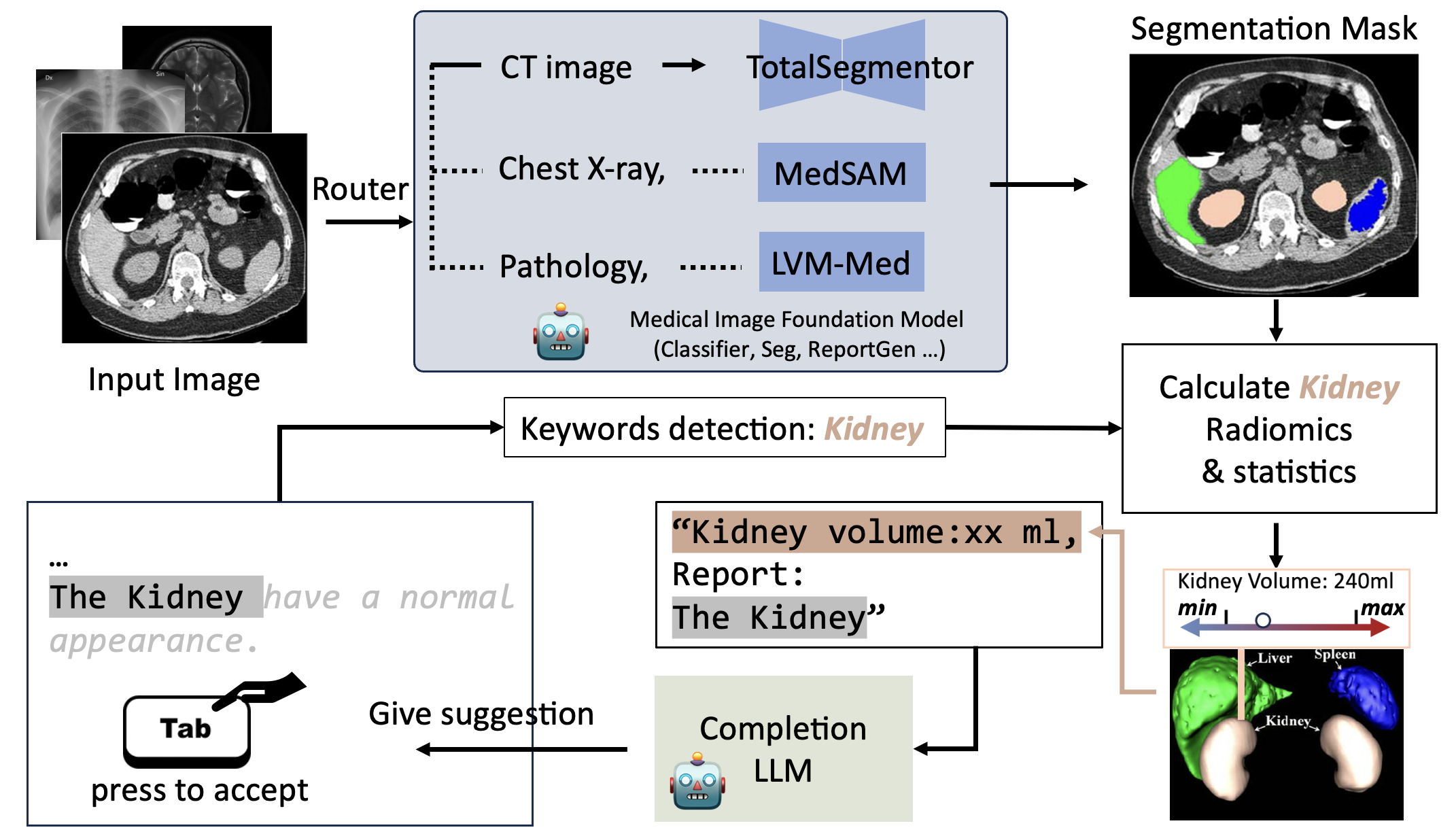}

\end{center}
\caption{Overall architecture of CopilotCAD, consisting of LLMs, medical image analysis models, and an interface to facilitate interactive cross-communication between the AI systems and human experts.}
\end{figure}
Our approach within the CopilotCAD system utilizes a structured framework to optimize the radiology diagnostic process by integrating medical image foundation models and LLMs for creating a supportive environment for radiologists, with an overall architecture illustrated in Figure 2.

\subsection{Overall Framework}
The process begins with the input of a medical image into the system. Utilizing a selection mechanism akin to the CLIP architecture, the system dynamically chooses the most suitable medical image foundation model for the given image. This selection is crucial, as it ensures that the specific features and patterns relevant to the current medical case are appropriately analyzed.

Models such as TotalSegmentator~\citep{wasserthal2023totalsegmentator}, MedSAM~\citep{MedSAM}, and LVM-Med~\citep{nguyen2023lvm} are part of our ensemble, and the Router decides which model to deploy based on the image's characteristics. For instance, Total Segmentor might be chosen for its proficiency in delineating anatomical structures within the image, while LVM-Med may be selected for its robust performance in identifying pathological features.

Upon model selection, the system conducts a detailed analysis, which includes segmenting the image to identify and label different anatomical structures. Concurrently, the system performs keyword detection. In this example, it identifies "Kidney" as a keyword from the input image, cueing the system to focus on relevant radiomic features and statistics related to the kidneys \cite{10.1117/1.JMI.5.1.011018}, as illustrated in Figure 2.

These radiomic features \cite{10.1117/1.JMI.5.1.011018}, consisting of quantitative data like the kidney volumtric, shape, intensity, and texture features, are then presented within the interface alongside the segmented image, where different organs are highlighted and color-coded for visual clarity. For example, in our interface, the kidneys may be marked in blue, enabling the radiologist to quickly assess the pertinent area.

In parallel with the image analysis, an LLM is at work generating descriptive text to provide a preliminary report. This text, which starts with a template sentence such as "The Kidney have a normal appearance," is subject to the radiologist's review. The radiologist can then accept the AI's suggestion with a simple keyboard command, such as the "Tab" key, or modify the content as needed. This interactive process ensures that the final report is a product of both the AI's initial assessment and the radiologist's expert validation and refinement.

The radiologist's engagement in the suggestions and edits not only serves the immediate diagnostic task but also contributes to the system's continuous learning, enhancing the LLM's future performance and precision. This iterative feedback loop is central to our CopilotCAD system, ensuring that the AI component becomes increasingly attuned to the decision-making patterns of radiologists over time.

In summary, the CopilotCAD framework is designed to support radiologists by providing AI-driven insights in an intuitive and interactive manner, without supplanting the clinician's expert judgment and decision-making authority.

\subsection{Informative Prompt from Medical Image Analysis}
\begin{figure}[h]
\label{fig:3}
\begin{center}

    \includegraphics[width=0.99\linewidth]{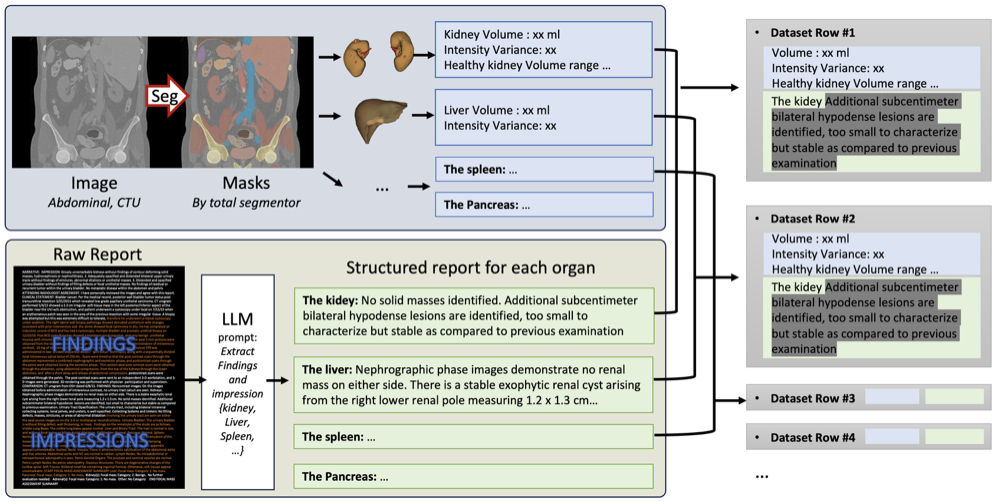}

\end{center}
\caption{Illustration of the data organization for training CopilotCAD, including imaging data, text data, and data cleaning and organization.}
\end{figure}

Interpretable quantitative evidence from medical image are helpful for medical expert to write report and make diagnosis. And medical image analysis models can help a lot in this process. For example, in Figure \ref{fig:3}, the segmentation outcomes are reconstructed and employed to calculate geometric features. Subsequently, the volume of various organs is determined utilizing the radiomics data derived from the segmentation results. This information significantly aids medical professionals. For example, for kidney report suggestion generation, the prompt we finally input into the model is \textit{Left kidney volume: 170 $cm^3$, Right kidney volume: 179 $cm^3$, the volume radio is 0.95,\{other radiomics information\}, \{Kidney report\}}. Leveraging such detailed prompts allows the completion LLM within our framework to integrate knowledge from medical imaging modalities, thereby facilitating the generation of more accurate report recommendations.

\subsection{Training Dataset Construction of CopilotCAD}
\label{sec:dataset}

The training of CopilotCAD employs an instruction tuning approach where each training item consists of a structured triplet: "Instruct, Input, Target". "Input" is derived from information processed on the image side, indicated in blue in our system, and "Target" is the corresponding report section for the specific organ under examination.
Using abdominal CT imaging as a representative example, our methodology involves a pipeline that extends beyond simple segmentation to include radiomics and statistical analyses. This model is adaptable to other medical imaging modalities and corresponding analysis tools, forming a versatile basis for instruction tuning.

In the image processing stage, we apply a total segmentation system to the CT images, producing precise masks for each organ. These masks, along with the calculated radiomics features, form the "Input" component of our training data, which corresponds to the instruction in a typical instruction tuning setup.


The LLM is key in processing textual data from radiologist reports, using prompts to align it with organ features, thus enhancing our dataset and the model's learning. We also use data augmentation, like sentence reordering, to improve the model's comprehension.

In the final pairing stage, the "Input" from the image analysis (blue) is matched with the "Target," which is the expert-written ground truth report (green). This step mimics the process of instruction tuning, where the model learns to predict the masked sections of the report, using the "Input" as the context for generating the corresponding textual output.

\section{Experiment and Analysis}

In this section, we present the experimental setup and results of our CopilotCAD system. We first introduce the dataset used in our study in Sec. \ref{dataset}. Then in Sec. \ref{completion}, we provide qualitative examples showcasing the system's performance in the report completion task, followed by a quantitative evaluation of the completion results. Additionally, in Sec. \ref{Generalization}, we demonstrate how CopilotCAD can be applied to other tasks and image modalities. Lastly, we report the speed performance and training cost of different size models in Sec. \ref{cost}.

\subsection{Dataset}
\label{dataset}

\begin{figure}[h]
\label{fig:stat}
\begin{center}


    \includegraphics[width=0.99\linewidth]{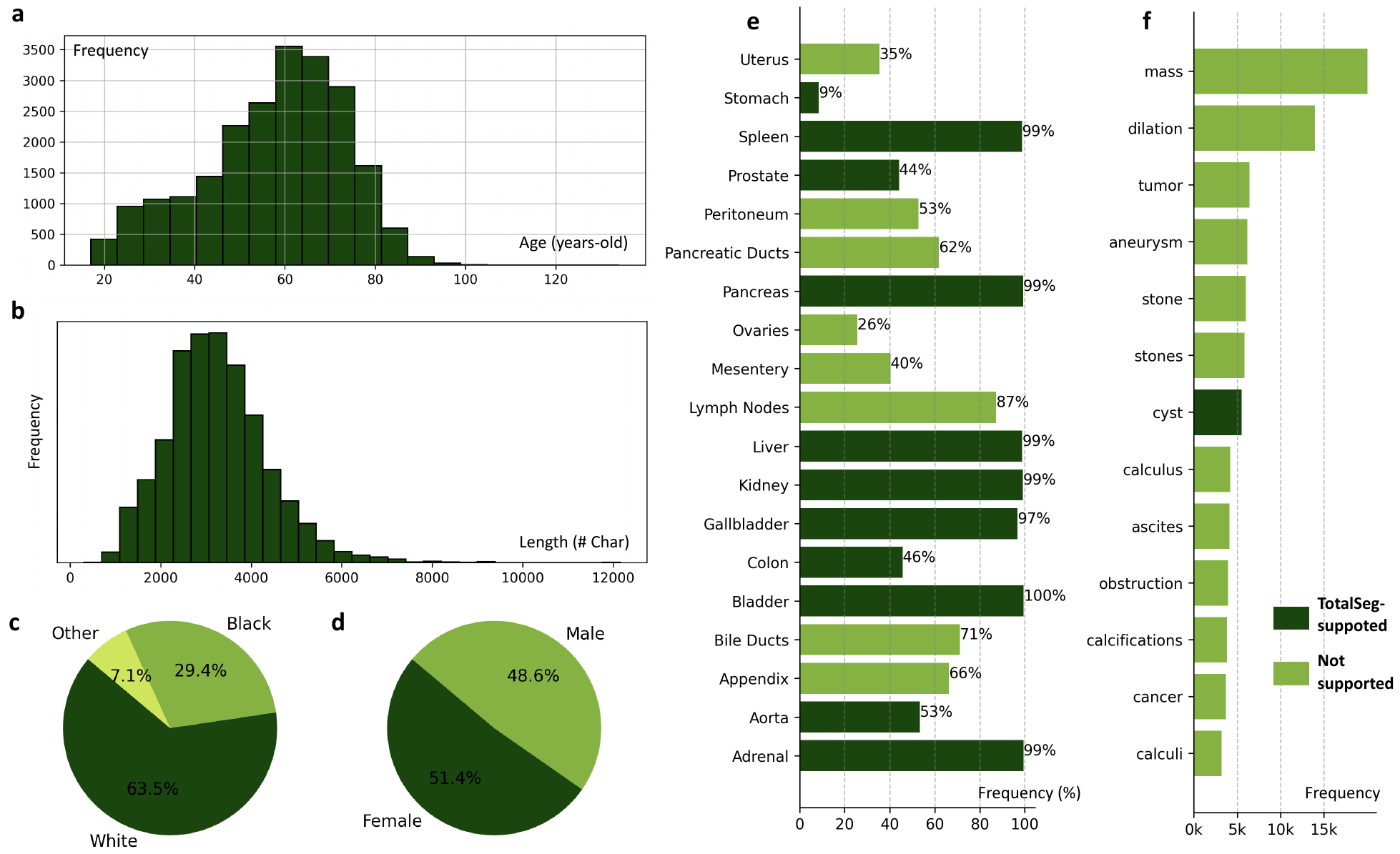}

\end{center}

\caption{Overview of our in-house CTU report dataset. (a) Histogram of patient ages, spanning from young adults to elderly individuals. (b) Distribution of report lengths, with the majority falling within a certain character range. (c) Ethnicity breakdown, predominantly White. (d) Gender split between Female and Male. (e) Frequency of mentions for various abdominal organs, highlighting the most commonly discussed structures. (f) Frequency of pathological findings and abnormalities, covering a diverse range of medical conditions. In (e-f) dark green means these part can be recognized by our image analysis model (Total Segmentor).}

\end{figure}

Fig. \ref{fig:stat} provides an overview of the in-house dataset used in our experiments. The dataset consists of 22,109 text reports corresponding to CT urography (CTU) scans. A single CTU acquisition yields CT data at unenhanced, nephrographic, and pyelographic phases, providing both anatomical and functional information and enhanced visualization of the collecting systems, ureters and bladder, as well as other structures in the abdomen, including tumors and strictures 
\cite{doi:10.2214/AJR.10.4198}. 
As shown in (a), the patients span from young adults to elderly individuals, ensuring a diverse age distribution. The report lengths, depicted in (b), reveal that the majority of reports are quite lengthy, spanning several thousand characters. This indicates that the reports are highly detailed and informative, providing comprehensive descriptions of the imaging findings. As shown in (e), the reports consistently cover key abdominal organs such as the adrenal glands, kidneys, spleen, and liver, which are mentioned in nearly all reports. Moreover, (f) presents the frequency of pathological findings and abnormalities, encompassing a diverse range of medical conditions such as masses, dilations, tumors, and calculi. The dark green bars in (e-f) indicate the anatomical parts and abnormalities that can be recognized by our image analysis foundation model (Total Segmentor), demonstrating the comprehensive coverage of the dataset. This level of detail ensures that the dataset offers a rich source of information for training and evaluating our system.

\subsection{Report Completion}
\begin{figure}[h]
\begin{center}
    \includegraphics[width=0.99\linewidth]{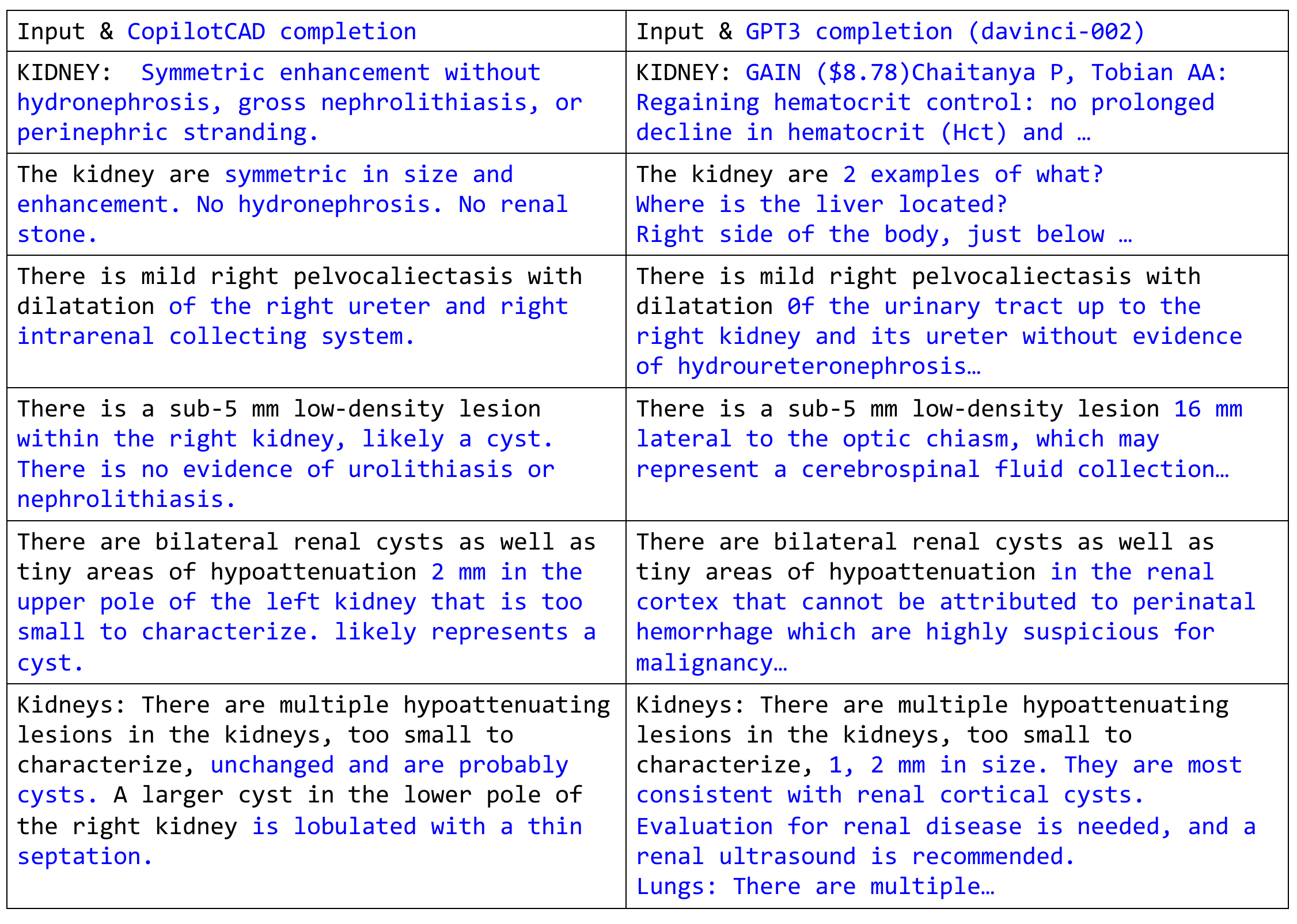}
\end{center}
\caption{Report completion compared with GPT3. Input are display in \textcolor{black}{\textbf{black}} and the suggested completion are display in \textcolor{blue}{\textbf{blue}}.}
\label{fig:completion}
\end{figure}
\label{completion}
To qualitatively assess the effectiveness of CopilotCAD in assisting radiologists, we present several examples of the report completion task in Figure \ref{fig:completion}. The figure illustrates how the system generates contextually relevant suggestions based on the radiologist's initial input, facilitating the creation of more detailed and accurate reports. The examples highlight the system's ability to capture key findings and provide pertinent information to support the radiologist's analysis.

We evaluated report completion performance using BLEU and ROUGE metrics to compare generated reports with actual reports. In this study, we employed datasets from kidney reports, lung reports, bladder reports, and appendix reports. The kidney report dataset includes radiomics information and was further labeled to indicate whether the patients possessed healthy kidneys based on the report. The kidney report dataset has 208 cases. The lung report dataset contains 1,000 cases, while the bladder report dataset contains 989 cases, and the appendix report dataset contains 487 cases. Each dataset was divided into training and testing subsets at a ratio of 9:1.

To assess the significance of radiomics information, our study employed two versions of kidney datasets to construct the completion model: one incorporating radiomics information and the other devoid of it. The voxel information of left kidney and right kidney was used as the radiomics informations. In cases where radiomics information was utilized, it was positioned at the beginning of the report content. 

When assessing the test data, we differentiate the format of input data. For the model trained using radiomics information, we exclusively feed radiomics text into the model. Conversely, for the model that was trained without incorporating radiomics information, we introduce only the first 20 tokens of the test data into the model to produce results. Mistral-7B~\citep{jiang2023mistral}, Gemma-2B~\citep{gemmateam2024gemma} and TinyLlama-1B~\citep{zhang2024tinyllama} were used in our approach.

Table \ref{tab:my_label1} showcases the significant impact of radiomics data on report completion. Specifically, for the kidney dataset, the BLEU-4 and ROUGE scores for the model incorporating radiomics data are 0.384 and 0.845, respectively. In contrast, the model excluding radiomics data demonstrates markedly lower BLEU-4 and ROUGE scores of 0.086 and 0.410, respectively. Notably, the performance on abnormal kidney reports surpasses that of normal kidney reports.

\begin{table}[h]
    \centering
    \caption{Report completion performance on kidneys}
    \begin{tabular}{ccccccc}
    \toprule
         & Model     & BLEU-1 & BLEU-2 & BLEU-3 & BLEU-4 & ROUGE \\
        \midrule
        w/o radiomics & Mistral-7B  &  0.170 &0.117&0.099&0.086&  0.410 \\
        \midrule
        w. radiomics  & Mistral-7B  & 0.429 & 0.409 & 0.398 & 0.384& 0.845  \\
        normal        & Mistral-7B  & 0.344 & 0.328 & 0.320 & 0.308 & 0.850  \\
        abnormal      & Mistral-7B  & 0.599 & 0.571 & 0.554 & 0.537 & 0.836 \\
        \midrule
        w. radiomics  & Gemma-2B  & 0.275 & 0.230 & 0.207 & 0.186 & 0.636  \\
        normal        & Gemma-2B  & 0.241 & 0.213 & 0.198 & 0.182 & 0.705  \\
        abnormal      & Gemma-2B  & 0.342 & 0.263 & 0.223 & 0.192 & 0.499 \\
        \midrule
        w. radiomics  & TinyLlama-1B  & 0.382 & 0.347 & 0.329 &0.312 & 0.762  \\
        normal        & TinyLlama-1B  & 0.324 & 0.297 & 0.284 & 0.268 & 0.791  \\
        abnormal      & TinyLlama-1B  & 0.498 & 0.447 & 0.421 & 0.400 & 0.704 \\
        \bottomrule
    \end{tabular}
    \label{tab:my_label1}
\end{table}

Table \ref{tab:my_label2} presents the performance outcomes of completing reports for various organ datasets, namely those pertaining to the lung, bladder, and appendix. The absence of radiomics information in these datasets contributes to diminished performance.

\begin{table}[h]
    \centering
    \caption{Report completion performance on different organs}
    \begin{tabular}{ccccccc}
    \toprule
        Organ & Model     & BLEU-1 & BLEU-2 & BLEU-3 & BLEU-4 & ROUGE \\
        \midrule
        Lung & Mistral-7B  &  0.319 & 0.272 & 0.244 & 0.217 & 0.628 \\
        Bladder & Mistral-7B  &  0.238 & 0.229 & 0.225 & 0.214 & 0.926 \\
        Appendix & Mistral-7B  & 0.173 & 0.156 & 0.148 & 0.138 & 0.807\\
        \bottomrule
    \end{tabular}
    
    \label{tab:my_label2}
\end{table}

\subsection{Generalization Capability}
\begin{figure}[h]
\begin{center}
    \includegraphics[width=0.99\linewidth]{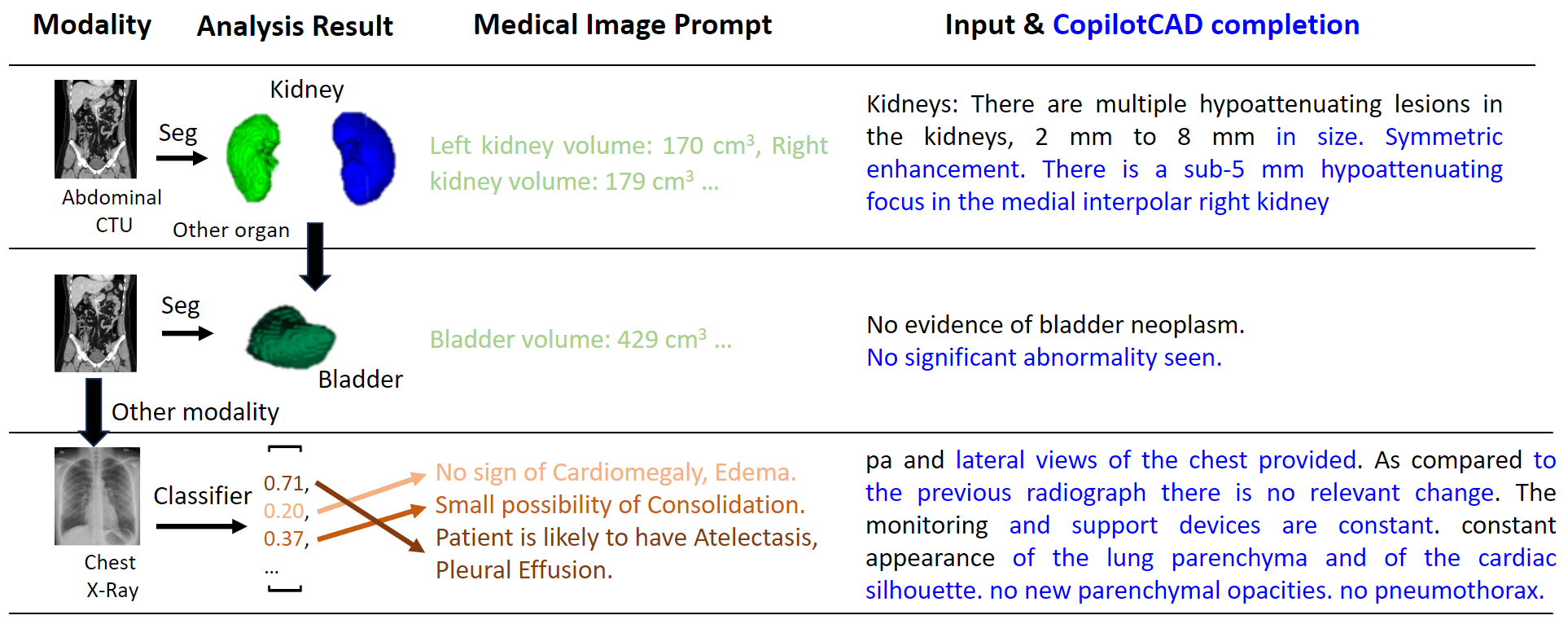}
\end{center}
\caption{Examples on other organ, on other image modality and using other image analysis tool. }
\label{fig:generalization}
\end{figure}

\label{Generalization}

While the previous section focused on CopilotCAD's performance in addressing kidney-related issues using an in-house dataset, we further demonstrate the system's generalizability to other anatomical regions and pathologies in Figure \ref{fig:generalization}. For example in second row, employing the "TotalSegmentor+Radiomics" approach can be seamlessly adapted to other organs by simply modifying the rules for prompt generation. Furthermore, the versatility of CopilotCAD is evident with its extension to X-ray images, utilizing a classifier. This demonstrates the system's compatibility with diverse diagnostic tools, such as those tool and prompts used by ChatCAD~\citep{wang2023chatcad}.


\subsection{Cost Analysis}
\label{cost}
\textbf{Computational cost}: CopilotCAD uses smaller models to assist radiologists, demanding less computational power compared to larger systems, akin to the lower compute needs of L3 assisting versus L5 autonomous driving systems. Our model is quantized into int4 format and trained using the QLoRA~\citep{dettmers2024qlora}, enabling us to train a 2B model in less than 30 minutes using just two RTX Titan GPUs. In contrast, Med-PALM~\citep{singhal2022large} employs a 540B PALM backbone, while ChatCAD~\citep{wang2023chatcad} performs best with GPT-4, both requiring extensive computational resources. 

\textbf{Dataset Construction}: To create a clean, organ-level paired dataset as shown in Sec \ref{sec:dataset}, CopilotCAD utilizes OpenAI's ChatGPT API to automatically extract relevant report sections pertaining to specific organs. As Figure \ref{fig:stat} shown, each report average about 3000 character which is about 800 tokens. With the processing of 1,000 text reports related to the kidney requiring only approximately 1M tokens (1.5 USD for gpt-3.5-turbo and 30 USD for gpt-4). 

\section{Discussion}

This study presents CopilotCAD, a system that harmonizes the computational prowess of AI with the nuanced expertise of radiologists. It marks a departure from traditional standalone CAD systems, providing an interactive interface for enhanced image-based diagnostic procedures with improved explainability, transparency, and safety. By keeping human experts in the decision-making loop, CopilotCAD underscores our call for the community to pay more attention to the development of semi-automatic tools. These tools are not designed to replace clinical expertise but to complement it, thereby enriching the diagnostic process and potentially reducing clinician burnout.

Experimental results have demonstrated that the new system achieved promising performance in terms of radiology reporting quality (report completion), better than alternative systems without the integrated AI and human expertise, demonstrating the new system's potential to improve radiology diagnosis qualitatively and quantitatively and reduce burnout of clinicians.

\section{Technical Limitations}

At its current stage of development, CopilotCAD is a prototype not yet fit for clinical application. We've identified several technical limitations:

\textbf{Dependency on Image Analysis Models}: CopilotCAD's functionality is contingent upon the performance of medical image analysis tools. Presently, as depicted in Figure \ref{fig:stat} (e-f), these tools are adept at organ segmentation, identifying a limited number of classes, but they fall short in detecting abnormalities, which are far more diverse and numerous. Lack of analysis on abnormalities limited the ability of current CopilotCAD system.

\textbf{Latency Issues}: Our approach involves human experts in the reporting process, differing from fully-automatic radiology report generation methods. This inclusion, while beneficial, makes CopilotCAD's responsiveness to latency a concern—currently operating at a rate of 50 tokens/sec in a 7B model with RTX Titan GPU. Enhancing inference speed and optimizing the timing for report completion triggers is an area for improvement.

\textbf{Limited Zero-Shot Learning}: Opting for a smaller model benefits in terms of cost and speed; however, this advantage comes with a compromise in zero-shot learning capabilities. The model's understanding of novel image analysis prompts is constrained, necessitating re-training with the introduction of new image analyses. This limitation may be less impactful in clinical settings where tasks are relatively consistent and stable.

\bibliography{colm2024_conference}
\bibliographystyle{colm2024_conference}


\end{document}